\documentclass[10pt,journal,compsoc]{IEEEtran}

\usepackage{tikz}
\usepackage{amsmath,amssymb} % define this before the line numbering.
\usepackage{color}
\usepackage{epsfig}
\usepackage{graphicx}

\usepackage{caption}
\usepackage{subcaption}

\usepackage{microtype}
\usepackage{booktabs}
\usepackage{enumitem}
\usepackage{array}
\usepackage{float}
\usepackage{soul}
\usepackage{comment}
\usepackage[ruled,vlined]{algorithm2e}
\usepackage{listings}
\usepackage{multirow}
\usepackage{booktabs}
\usepackage{adjustbox}

\definecolor{mydarkblue}{rgb}{0.0,0.0,0.9}

\usepackage[normalem]{ulem}
\useunder{\uline}{\ul}{}

\usepackage[accsupp]{axessibility}  % Improves PDF readability for those with disabilities.

\usepackage[pagebackref,breaklinks,colorlinks]{hyperref}

\begin{document}

\title{RFNet-4D++: Joint Object Reconstruction and Flow Estimation from 4D Point Clouds with Cross-Attention Spatio-Temporal Features}

\author{Tuan-Anh~Vu,~\IEEEmembership{Graduate Student Member,~IEEE,}
        Duc~Thanh~Nguyen,~\IEEEmembership{Member,~IEEE,}
        Binh-Son~Hua,
        Quang-Hieu~Pham,
        and~Sai-Kit~Yeung,~\IEEEmembership{Senior Member,~IEEE}% <-this % stops a space
\IEEEcompsocitemizethanks{\IEEEcompsocthanksitem T.A. Vu, and S.K. Yeung are with School of Computer Science and Engineering, The Hong Kong University of Science and Technology, Hong Kong SAR. 
\IEEEcompsocthanksitem S.K. Yeung is also with Division of Integrative Systems and Design, The Hong Kong University of Science and Technology, Hong Kong SAR.
\IEEEcompsocthanksitem D.T. Nguyen is with School of Information Technology, Deakin University, Australia.
\IEEEcompsocthanksitem B.S. Hua is the School of Computer Science and Statistic, Trinity College Dublin, Ireland.
\IEEEcompsocthanksitem Q.H. Pham is with Woven by Toyota, USA.
\IEEEcompsocthanksitem Corresponding author: T.A. Vu. }% <-this % stops an unwanted space
% \thanks{Manuscript received April 19, 2005; revised August 26, 2015.}
}

\markboth{IEEE Transactions on Pattern Analysis and Machine Intelligence}%
{Vu \MakeLowercase{\textit{et al.}}: RFNet-4D++: Joint Object Reconstruction and Flow Estimation from 4D Point Clouds with Cross-Attention Spatio-Temporal Features}

\IEEEtitleabstractindextext{%
\begin{abstract}
Object reconstruction from 3D point clouds has been a long-standing research problem in computer vision and computer graphics, and achieved impressive progress. However, reconstruction from time-varying point clouds (a.k.a. 4D point clouds) is generally overlooked. In this paper, we propose a new network architecture, namely RFNet-4D++, that jointly reconstructs objects and their motion flows from 4D point clouds. The key insight is simultaneously performing both tasks via learning of spatial and temporal features from a sequence of point clouds can leverage individual tasks, leading to improved overall performance. To prove this ability, we design a temporal vector field learning module using an unsupervised learning approach for flow estimation task, leveraged by supervised learning of spatial structures for object reconstruction. Extensive experiments and analyses on benchmark datasets validated the effectiveness and efficiency of our method. As shown in experimental results, our method achieves state-of-the-art performance on both flow estimation and object reconstruction while performing much faster than existing methods in both training and inference. Our code and data are available at \url{https://github.com/hkust-vgd/RFNet-4D}.
\end{abstract}

}

\maketitle

\IEEEdisplaynontitleabstractindextext

\IEEEpeerreviewmaketitle

\IEEEraisesectionheading{\section{Introduction}\label{sec:introduction}}

\begin{figure}[t]
\centering
\includegraphics[width=0.9\linewidth]{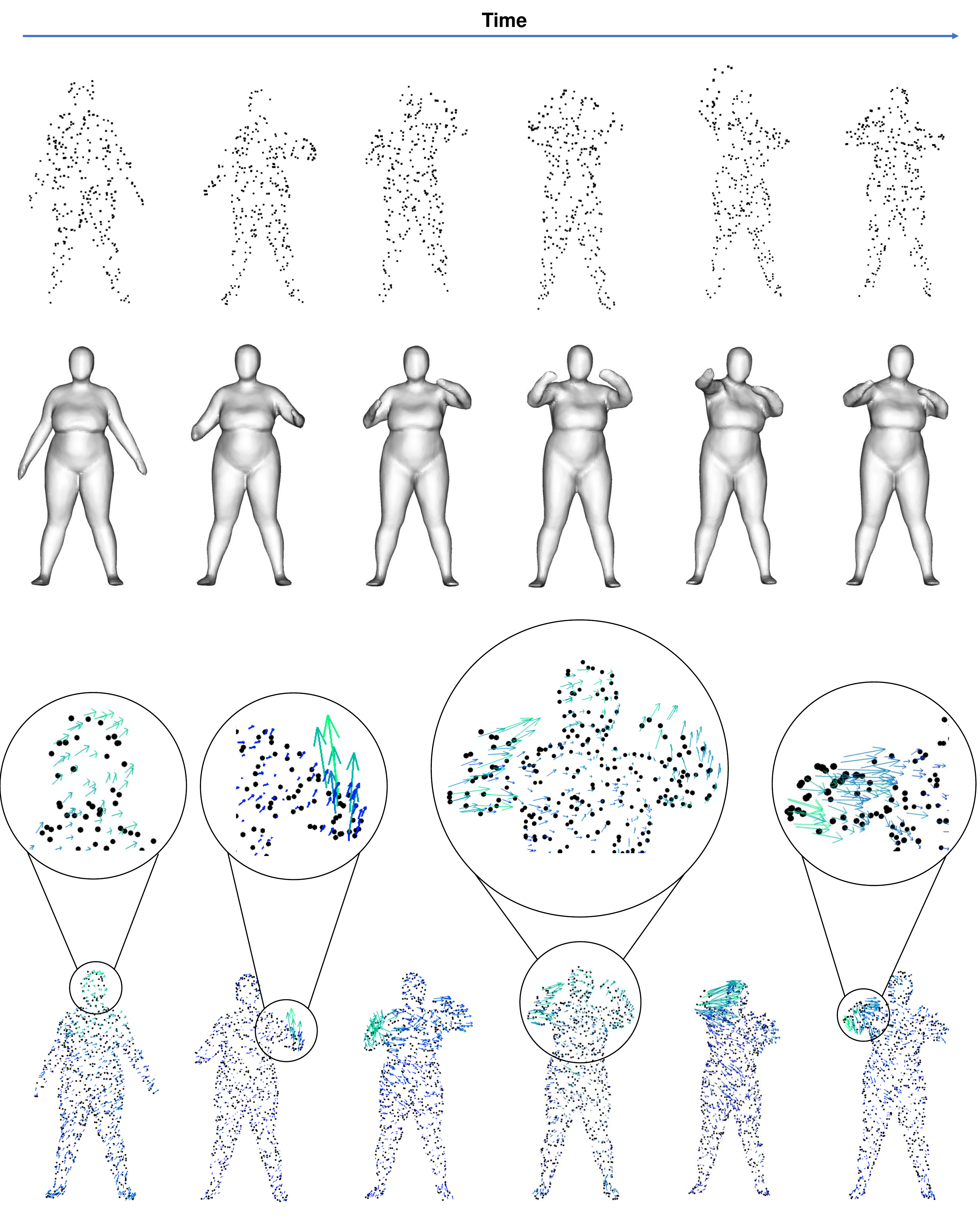}
\caption{\textbf{Summary of our method.} Given a sequence of time-varying 3D point clouds (first row), we jointly reconstruct corresponding 3D geometric shapes (second row) and estimate their motion fields for every point cloud (third row).}
\label{fig:teaser}
\end{figure}

\IEEEPARstart{L}{iterature} has shown several breakthroughs in deep learning for reconstruction of 3D models from point clouds. Recently, the research community has seen great successes in neural representations using implicit fields~\cite{mescheder2019,park2019,chen2019,Michal2019}, which pave an effective way on how 3D data can be represented by neural networks. Unlike traditional representations that are often realized in discrete forms (e.g., discrete grids of pixels in image representation, discrete grids of voxels in 3D object representation), the neural implicit representation parameterizes a signal as a continuous function via a neural network. This function maps a signal from its original domain, which can be queried at any resolution, to an output domain that captures some properties of the query. Most existing methods focus on neural representation of 3D data in static conditions. However, in reality, real-world objects exist in a dynamic environment that changes over time and space, and thus cannot be well modeled using implicit representations applied to static shapes. Approaches for 4D reconstruction (i.e., reconstruction of a 3D object over time) have been explored but they often need expensive multi-view settings~\cite{Leroy2017,Coskun2017,Mustafa2015,Mustafa2016}. These settings rely on a template model (of the object) with fixed topology~\cite{alldieck2018,Kanazawa2019,Tung2017,Zheng2017}, or require smooth spatio-temporal input~\cite{Pekelny2008,Michael2007}, and thus limiting their applicability in practice.

To enable object reconstruction directly from 4D data without predefined templates, OFlow~\cite{Niemeyer2019}, a pioneering method for 4D reconstruction, was developed to calculate motion fields of 3D points in a 3D point cloud in space and time to implicitly represent trajectories of all the points in dense correspondences between occupancy fields. To learn the motion fields in both space and time domains, OFlow made use of a spatial encoder to learn the spatial structure of the input point cloud and a temporal encoder to learn the temporal changes of the point cloud in time. Despite impressive reconstruction results, this paradigm has a number of drawbacks. First, its spatial encoder does not take geometric attributes from numerous frames into consideration, impairing the capacity to precisely reconstruct geometric surfaces. Neither does its temporal encoder take into account temporal correspondences, which are critical for accurately capturing temporal dynamics. Second, errors in prediction of temporal continuity and reconstructed geometries are accumulated by integral of estimated instantaneous findings. Third, OFlow is trained using supervised learning. This requires correspondence labelling for all 3D points across frames in training data, leading to high labelling costs and low scalability. Fourth, the method exhibits low computational efficiency in both training and inference phases. This is due to expensive computations required to sequentially determine trajectories of 3D points throughout time by solving complex ordinary differential equations.

To address the aforementioned challenges, we propose a network architecture, namely RFNet-4D++, for 4D reconstruction and flow estimation of dynamic point clouds. Our key idea is to jointly perform two tasks: 4D reconstruction and flow estimation with an intention that each task can leverage the other one to improve the overall performance. Specifically, our network takes a sequence of 3D point clouds of an object over time as input, then encodes the point clouds into spatio-temporal representations using a compositional encoder. These spatio-temporal representations are formed inclusively. In particular, the spatio-temporal representation of a point cloud at a time step is calculated from the spatial layout of points in that point cloud and the temporal changes of the points in the point cloud throughout time. The spatio-temporal representations are then decoded by a joint decoder which jointly reconstructs the object and predicts a motion vector for each point in the reconstructed object throughout time. The entire network can be trained end-to-end, where the reconstruction and flow estimation tasks are trained with supervised and unsupervised learning, respectively. Our method also allows fast computations of spatial and temporal features as those computations can be performed in parallel. This is another advantage of our method, compared with OFlow which estimates the motion flows sequentially and thus often experiences time lags. We illustrate several results of our method in Fig.~\ref{fig:teaser}. In summary, the contributions of our work are as follows:
\begin{itemize}
    \item RFNet-4D++: a network architecture for joint object reconstruction and flow estimation from a sequence of time-varying 3D point clouds.
    \item A joint learning method for training the proposed RFNet-4D++ using both supervised and unsupervised learning, and in both forward and backward time direction. To the best of our knowledge, this learning mechanism is novel, and its benefit is verified throughout experiments.
    \item Extensive experiments and analyses that prove the effectiveness and efficiency of our proposed method on two tasks: 4D reconstruction and flow estimation.
\end{itemize}

\noindent A preliminary version of this work (RFNet-4D) has been published in~\cite{tavu2022rfnet4d}. RFNet-4D++ extends RFNet-4D in the following aspects:
\begin{itemize}
    \item We propose a new spatio-temporal representation that utilizes dual cross-attention mechanism to improve the performance of our method in both object reconstruction and flow estimation. 
    \item We extend the analysis of our method and its main components in various aspects. In particular, we investigate the impact of the spatial and temporal resolutions in input data to the performance of our method. We validate the main components of our method in a related work (LPDC~\cite{tang2021}).
    \item We further evaluate our proposed method on DeformingThing4D~\cite{li20214dcomplete}, a diverged and large-scale dataset of non-rigid objects ranging from humanoids to various animal species.
\end{itemize}
We show that these extensions further strengthen our method which achieves state-of-the-art results on both D-FAUST and DeformingThing4D dataset. 

%%%%%%%%%%%%%%%%%%%%%%%%%%%%%% Related %%%%%%%%%%%%%%%%%%%%%%%%%%%%%%

\section{Related Work}
\label{sec:related_work}

\subsection{3D Reconstruction} 
Numerous studies have been conducted with the goal of reconstructing a continuous surface from a variety of inputs, including RGB images~\cite{Tatarchenko2017,Wen2019,Kato2018}, point clouds~\cite{Kazhdan2006}. Thanks to advances in deep learning, recent 3D object reconstruction approaches have resulted in significant progress. Early attempts represent reconstructed objects in regular grid of 3D voxels~\cite{Wang2017ocnn,Girdhar16} or point clouds~\cite{qi2018frustum,Fan2017}. However, those representations cannot well capture surface details and suffer from low resolutions. Alternatively, there are methods, e.g., \cite{wang2018pixel,Liao2018,Kanazawa2018} reconstructing triangular meshes (including vertices and faces) of 3D objects. In these methods, an initial template with fixed topology is employed and the reconstruction is performed using regression. For surface representation, several methods focus on learning an implicit field function that allows more variable topology in reconstructed objects~\cite{Chibane2020,chabra2020,Erler2020,Chiyu2020}.

To extend the ability of implicit functions on representations other than traditional forms (i.e., voxels, points, meshes), occupancy maps~\cite{mescheder2019,peng2020} and distance fields~\cite{park2019,chabra2020} are proposed. An occupancy map of a 3D point cloud contains indicators that indicate being foreground of points in the 3D space. A distance field provides the distance from every point to its nearest surface. Since the implicit function models objects in a continuous manner, more information is preserved and more complicated shapes can be well described. For instance, Occupancy Network in~\cite{mescheder2019} described a 3D object using continuous indicator functions that indicate which sub-sets of 3D space the object occupies, and an iso-surface retrieved by employing the Marching Cube algorithm~\cite{marchingcubes}.

\subsection{4D Reconstruction} 
Despite being less studied compared with 3D reconstruction, literature has also shown recent attention of the research community to 4D reconstruction, i.e., reconstruction of a sequence of 3D objects from time-varying point clouds~\cite{Leroy2017,Mustafa2015,Mustafa2016}. In this section, we limit our review to only learning-based 4D reconstruction methods.

A crucial component in 4D reconstruction is motion capture and modelling. Niemeyer \textit{et al.}~\cite{Niemeyer2019} introduced a learning-based framework that calculates the integral of a motion field specified in space and time to implicitly represent the trajectory of a 3D point to generate dense correspondences between occupancy fields. Jiang \textit{et al.}~\cite{jiang2021} proposed a compositional representation for 4D capture, i.e., a deformable representation that encloses a 3D shape and velocity of its 3D points over time. Such representation was composed of encoder-decoder architectures. Specifically, to simulate the motion in time-varying 3D data, a neural Ordinary Differential Equation was trained to update the starting state of motion based on a learned motion representation, and a decoder was used to reconstruct a 3D model at each time step using a shape representation and the updated state. They also introduced an Identity Exchange Training technique to motivate their system to learn how to decouple each encoder-decoder successfully. Tang \textit{et al.}~\cite{tang2021} proposed a pipeline for determining the temporal evolution of the 3D shape of the human body using spatially continuous transformation functions between cross-frame occupancy fields. By explicitly learning continuous displacement motion fields from spatio-temporal shape representations, the pipeline aims to construct dense correspondences between projected occupancy fields at different time steps.

\subsection{Motion Transfer} 
Traditional techniques for 3D pose transfer often use discrete deformation transfers. An example is described in~\cite{wang2020neural}, where spatially adaptable instance normalisation~\cite{Huang2017} was used to modify 3D meshes. However, this method requires a dense triangular mesh of an object to be given in advance, while there is a specific mechanism to depict continuous flows in both spatial and temporal domains. 

3D motion transfer is another technique for creating 3D objects from a pair of source and target object sequences. It operates by causing the target object sequence to undergo the same temporal deformation in the source object sequence. This technique can be applied to model continuous transformation of an object's pose over time. For instance, OFlow~\cite{Niemeyer2019} transmitted motion across sequences of source and target human models by applying motion field-based representations to the targets in a predetermined manner. However, since OFlow does not explicitly differentiate the representations of pose and shape, we found that it only produces reasonable motion transfers when the identities and poses of the source and target objects are similar.

\subsection{Shape Correspondence Modelling} 
Modelling of point-to-point correspondences between two 3D shapes is a well-studied topic in computer vision and computer graphics~\cite{Biasotti2016}. Time-varying occupancy field learning is strongly related to field-based deformation~\cite{Marcel2018}, which we have previously discussed. However, most of these works describe the motion fields only on object surfaces. To better describe the motion flow, we argue to model the correspondences between the entire 3D shapes. 

When modelling the growth of a signed distance field, Miroslava \textit{et al.} ~\cite{Slavcheva2017} chose to implicitly provide the correspondences rather than explicitly yielding them. They optimized an energy function capturing the similarity between the Laplacian eigenfunction representations of the input and the target shape. However, we found that their method is sensitive to noise, probably due to lack of capability to provide correspondences accurately from signed distance fields. In contrast, we learn the rich correspondences between time-varying occupancy fields based on a intuitive insight, that the occupancy values of points are always invariant during the temporal evolution of the occupation field.

There are methods modelling deformable shapes overtime using implicit functions (continuous signed distance fields), e.g., DIT~\cite{zheng2021dit}, NDF~\cite{sun2022ndf}, ImplicitAtlas~\cite{yang2022implicit}. DIT~\cite{zheng2021dit} employed LSTM~\cite{sepp1997lstm} to model smooth deformations, while NDF\cite{sun2022ndf} utilized NODE~\cite{chen2018node} to achieve diffeomorphic deformations. On the other hand, ImplicitAtlas~\cite{yang2022implicit} leveraged the integration of multiple templates to enhance the capability of shape modelling. Remarkably, this improvement in shape modelling comes at a minimal computational cost, making ImplicitAtlas~\cite{yang2022implicit} an efficient choice for practical implementation. By combining multiple templates, ImplicitAtlas~\cite{yang2022implicit} is shown to effectively capture and represent complex shape variations while maintaining computational efficiency.

% \son{Maybe update the related work section with some new works.}

\begin{figure*}[t]
  \centering
  \includegraphics[width=\linewidth]{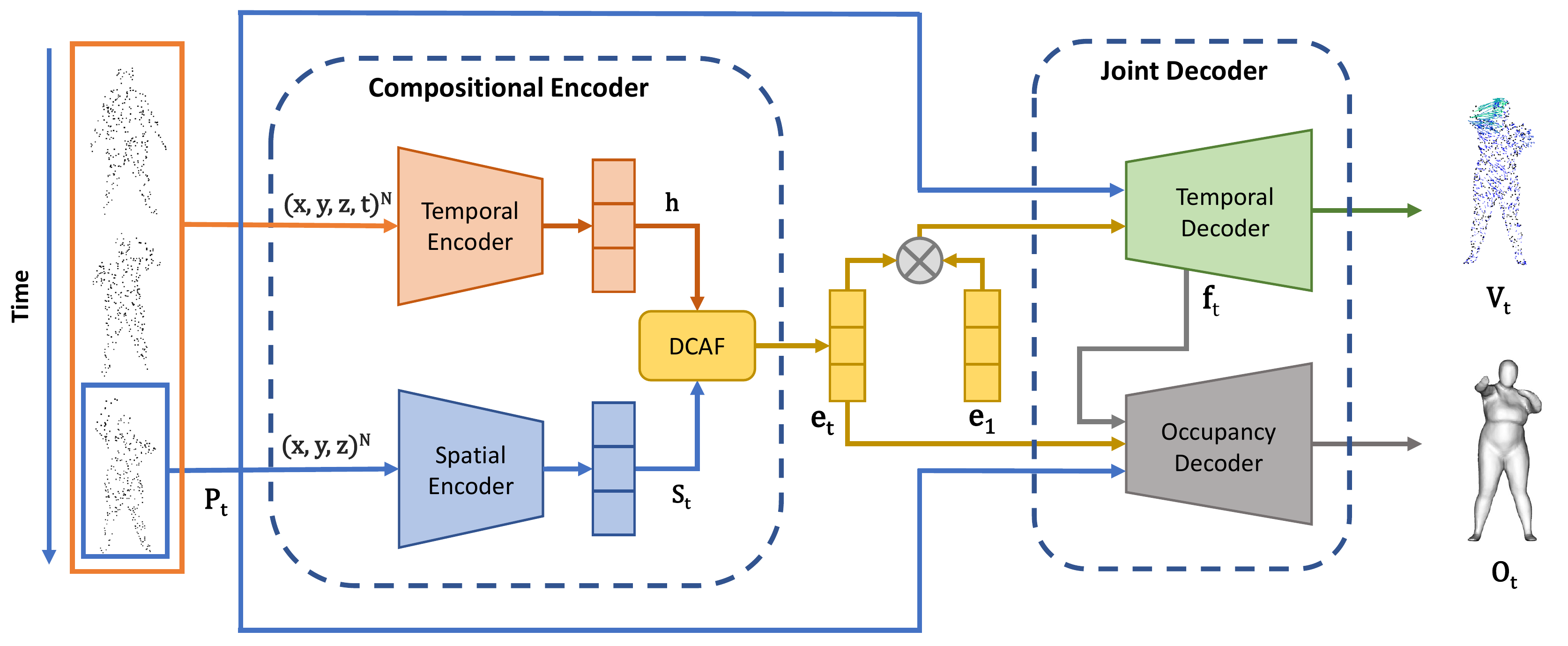}
  \caption{
  \textbf{Overview of our method.} An input 3D point cloud sequence is fed into a spatio-temporal encoder to extract spatio-temporal representations. The representations are then passed via two distinct decoders, occupancy and motion decoders. In each data frame, the occupancy decoder aims to predict an occupancy field of the point cloud in the frame. Simultaneously, the motion decoder predicts the correspondences between points in the current frame and its preceding frame. $\otimes$ indicates a concatenation operation.}
  \label{fig:pipeline}
\end{figure*}

\begin{figure*}[t]
  \centering
  \includegraphics[width=0.9\linewidth]{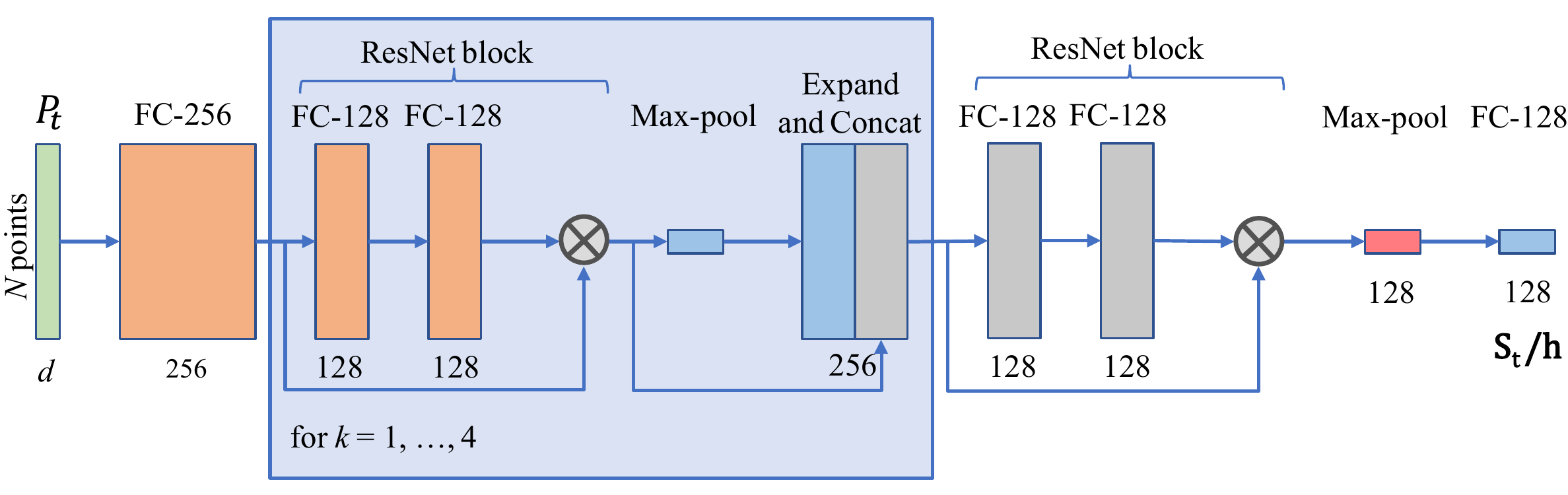}
  \caption{\textbf{Spatial/temporal encoder.} The input dimension $d$ is set accordingly to a corresponding encoder. In particular, $d=3$ (i.e., $(x, y, z)$-coordinates) for the spatial encoder and $d=4$ (i.e., $(x, y, z)$-coordinates and time variable) for the temporal encoder. $\otimes$ indicates a concatenation operation. Outputs of the spatial and temporal encoder are $\mathbf{S}_t$ and $\mathbf{h}$, respectively.}
  \label{fig:encoders}
\end{figure*}

%%%%%%%%%%%%%%%%%%%%%%%%%%%%%% Method %%%%%%%%%%%%%%%%%%%%%%%%%%%%%%

\section{RFNet-4D++}
\label{sec:proposed_method}

\subsection{Overview}

Our network takes as input a sequence of sparse, incomplete, and noisy 3D point clouds $\{P_t | t = 1,...,T\}$ where $T$ is the length of the sequence, and each point cloud $P_t$ is a set of 3D locations. Our aim is to simultaneously perform the following tasks:
\begin{itemize}
    \item Reconstruct a sequence of occupancy maps $\{O_t | t = 1,...,T\}$ where each $O_t$ is an occupancy map of a point cloud $P_t$, i.e., $O_t(\mathbf{p})=1$ if $\mathbf{p}$ is a 3D point on the reconstructed surface of $P_t$, and $O_t(\mathbf{p})=0$, otherwise;
    \item Estimate a sequence of vector fields $\{V_{t} | t=1,...,T\}$ where each $V_{t}$ is a 3D vector field capturing motion flows of reconstructed points of $P_t$, i.e., $V_t(\mathbf{p}) \in \mathbb{R}^3$ represents the motion flow of a reconstructed point $\mathbf{p}$ at time step $t$ given a point cloud $P_t$.
\end{itemize}

Both tasks benefit from a compositional encoder that learns spatio-temporal representations from time-varying point clouds. The temporal features contained in these spatio-temporal representations capture holistic motion information and are computed once on the entire input point cloud sequence. This allows fast computations in following operations as spatio-temporal data can be processed at any arbitrary frame. The spatio-temporal representations are  processed by a joint decoder including two decoders, each of which extracts relevant information for its downstream task (i.e., reconstruction and flow estimation). These decoders do not operate independently but cooperate closely to fulfill their tasks. To further exploit the benefit of temporal information, we couple the reconstruction and flow estimation tasks in both forward and backward time directions. We present an overview of our method in Fig.~\ref{fig:pipeline}. We describe the main components of our method in the following sections.

\begin{figure*}[t]
  \centering
  \includegraphics[width=0.92\linewidth]{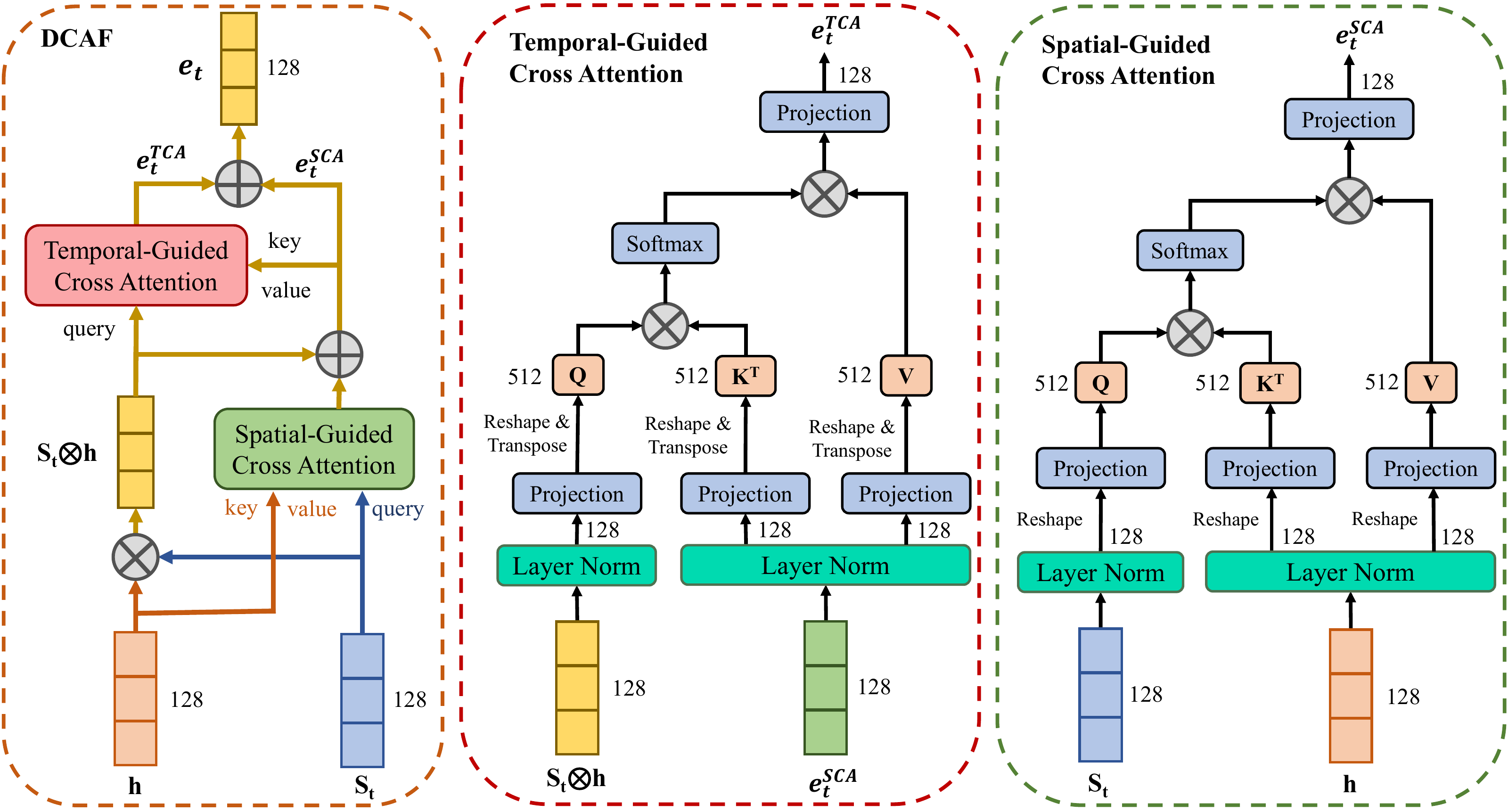}
  \caption{\textbf{Dual cross-attention fusion (DCAF) module and its sub-modules: Spatial-guided Cross Attention (SCA) and Temporal-guided Cross Attention (TCA).} Inputs are the spatial and temporal representations $\mathbf{S}_t$ and $\mathbf{h}$, and output is a spatio-temporal representation $\mathbf{e}_t$ which captures long-range contextual information in both temporal and spatial dimensions. $\oplus$ denotes an element-wise addition operation and $\otimes$ denotes a concatenation operation.}
  \label{fig:dcaf}
\end{figure*}

\subsection{Compositional Encoder} 
The compositional encoder includes a temporal encoder and a spatial encoder. There exist several manners to encode 4D point clouds. For instance, Liu \textit{et al.}~\cite{liu2019} applied spatio-temporal neighborhood queries in representing 4D point clouds. However, this method requires high computational complexity. Inspired by the success and efficiency of the point cloud representation used in OFlow~\cite{Niemeyer2019} and LPDC~\cite{tang2021}, we adopt the parallel ResNet~\cite{he2015deep} variant of PointNet~\cite{PointNet} for both the spatial and temporal encoder (see Fig.~\ref{fig:encoders}). These encoders are basically similar in their architectures. The difference between them is that while the spatial encoder processes each point cloud $P_t$ individually at a time $t$ to generate a representation $\mathbf{S}_t$, the temporal encoder acquires the whole point cloud sequence to calculate a holistic temporal representation $\mathbf{h}$ once.

The spatial and temporal representations are finally fused to form a spatio-temporal representation $\mathbf{e}_t$ that encodes the geometric information of a point cloud $P_t$ in space with regard to its temporal changes (see Fig.~\ref{fig:pipeline}). Our encoders share similar structures with the encoders in LPDC~\cite{tang2021}. In our original work (RFNet-4D)~\cite{tavu2022rfnet4d}, we created the spatio-temporal representation by simply concatenating the spatial and temporal features. However, such simple concatenation does not effectively capture the topology the point cloud's structure over time (e.g., different parts of a human body in motion can have different velocities while still obeying topological rules of deformation). To overcome this issue, we introduce here a dual cross-attention fusion module (DCAF) that effectively learns the correlation between spatial and temporal features. Our DCAF is inspired by the works in~\cite{jaegle2021,ding2022davit}, which learn attention scores across different modalities.  
% In the first cross-attention module, we utilize the spatial representation $\mathbf{S}_t$ for a query and the temporal representation $\mathbf{h}$ for a key and a value. This configuration enables us to effectively capture inter-point spatial dependencies in a point cloud, allowing for extraction of a global spatial context. The second cross-attention module takes two inputs: a concatenated representation (from $\mathbf{S}_t$ and $\mathbf{h}$) which serves as a query, and the output of the first cross-attention module which forms a key and a value. Unlike the first cross-attention module, the second cross-attention module aims to extract a global temporal context by integrating all spatial positions across different timestamps.

We describe the architecture of the DCAF module in Fig.~\ref{fig:dcaf}. The DCAF module includes two sub-modules: a Spatial-guided Cross Attention (SCA) module and a Temporal-guided Cross Attention (TCA) module. In the SCA module, we utilise the spatial representation $\mathbf{S}_t$ for a query and the temporal representation $\mathbf{h}$ for a key and a value. This configuration enables us to effectively capture inter-point spatial dependencies in a point cloud, allowing for extraction of a global spatial context $\mathbf{e}^{SCA}_t$. The TCA module takes two inputs: a concatenated representation (from $\mathbf{S}_t$ and $\mathbf{h}$) which serves as a query, and the output of the SCA module $\mathbf{e}^{SCA}_t$ which forms a key and a value. Unlike the SCA module, the TCA module aims to extract a global temporal context $\mathbf{e}^{TCA}_t$ by integrating all spatial positions across different timestamps. Each the SCA/TCA module consists of layer normalisation and projection implemented via fully connected layers. The spatio-temporal representation $\mathbf{e}_t$ is finally formed by element-wise addition of $\mathbf{e}^{SCA}_t$ and $\mathbf{e}^{TCA}_t$. 

Following the conventional definition of cross attention~\cite{jaegle2021,ding2022davit}, the output of a cross attention module (SCA/TCA) can be expressed as:
\begin{equation}
\label{eq:cross}
\text{Cross-Attention} (Q,K,V)=\text{Softmax}\left(\frac{QK^{\top}}{\sqrt{C}}\right) V
\end{equation}
where $Q,K,V$ represent the projected queries, keys and values, respectively, $\frac{1}{\sqrt{C}}$ is the scaling factor where $C$ is the number of total channels.

Depending on the particular cross attention module (SCA/TCA), $Q,K,V$ are defined accordingly. Specifically, as shown in Fig.~\ref{fig:dcaf}, for the SCA, $Q$ is calculated from $\mathbf{S}_t$, while $K$ and $V$ are derived from $\mathbf{h}$. On the other hand, for the TCA, $Q$ is obtained from the concatenation of $\mathbf{S}_t$ and $\mathbf{h}$, while $K$ and $V$ are determined from the output of the SCA.

Since $\mathbf{h}$ is computed once on the entire input point cloud sequence, $\mathbf{e}_t$ can be extracted at any arbitrary time step $t$ without time lags, as opposed to methods processing point clouds sequentially, e.g., OFlow~\cite{Niemeyer2019}. Thanks to this advantage, the processing time RFNet-4D++ can be optimized by calculating the spatio-temporal representations $\mathbf{e}_t$ for all the time steps $t$ in parallel.

\begin{figure*}[t]
    \centering
    \begin{subfigure}[b]{0.49\textwidth}
        \centering
        \includegraphics[width=\textwidth]{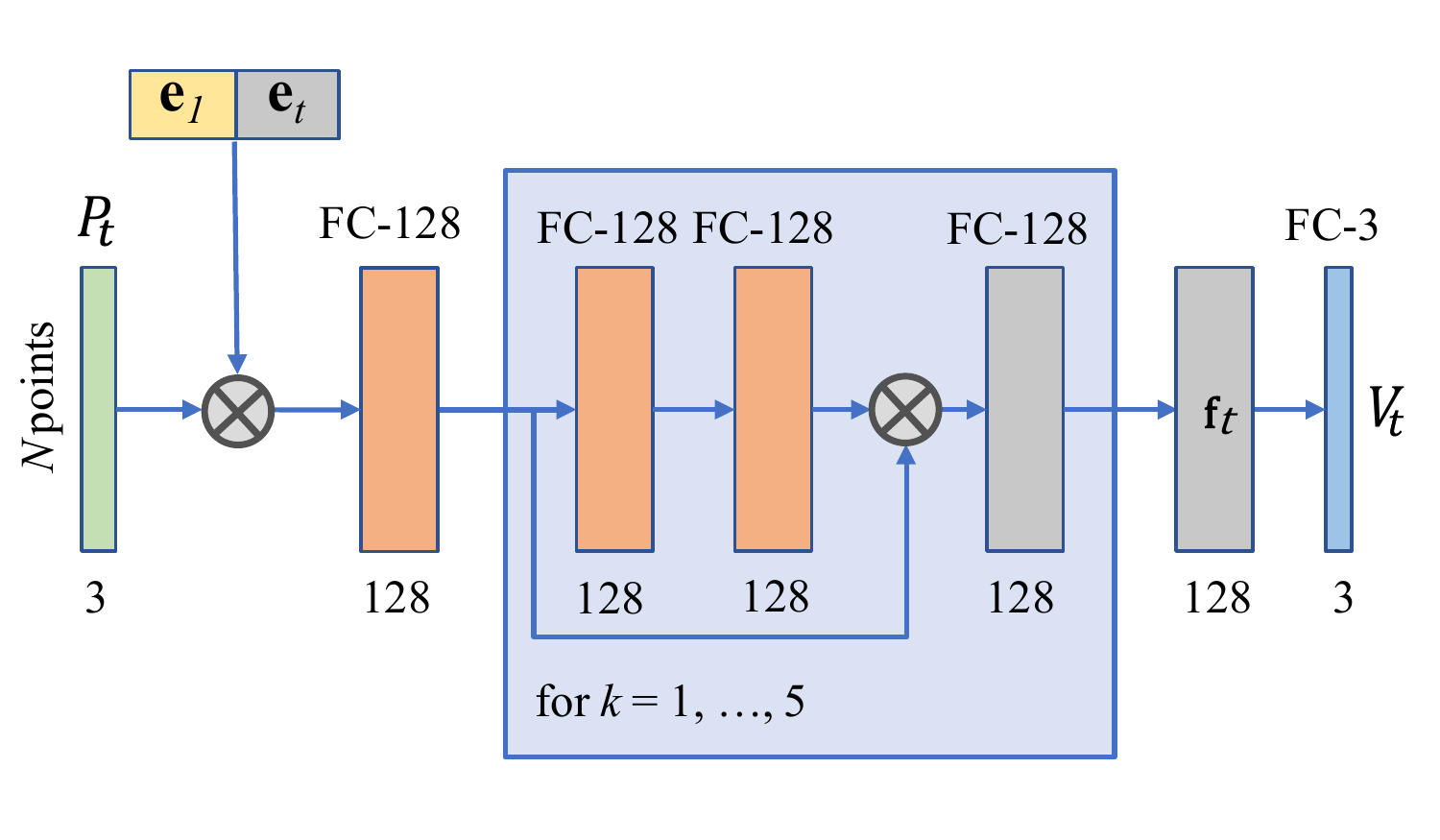}
        \caption{Temporal Decoder}
    \end{subfigure}
    \hfill
    \begin{subfigure}[b]{0.49\textwidth}
        \centering
        \includegraphics[width=\textwidth]{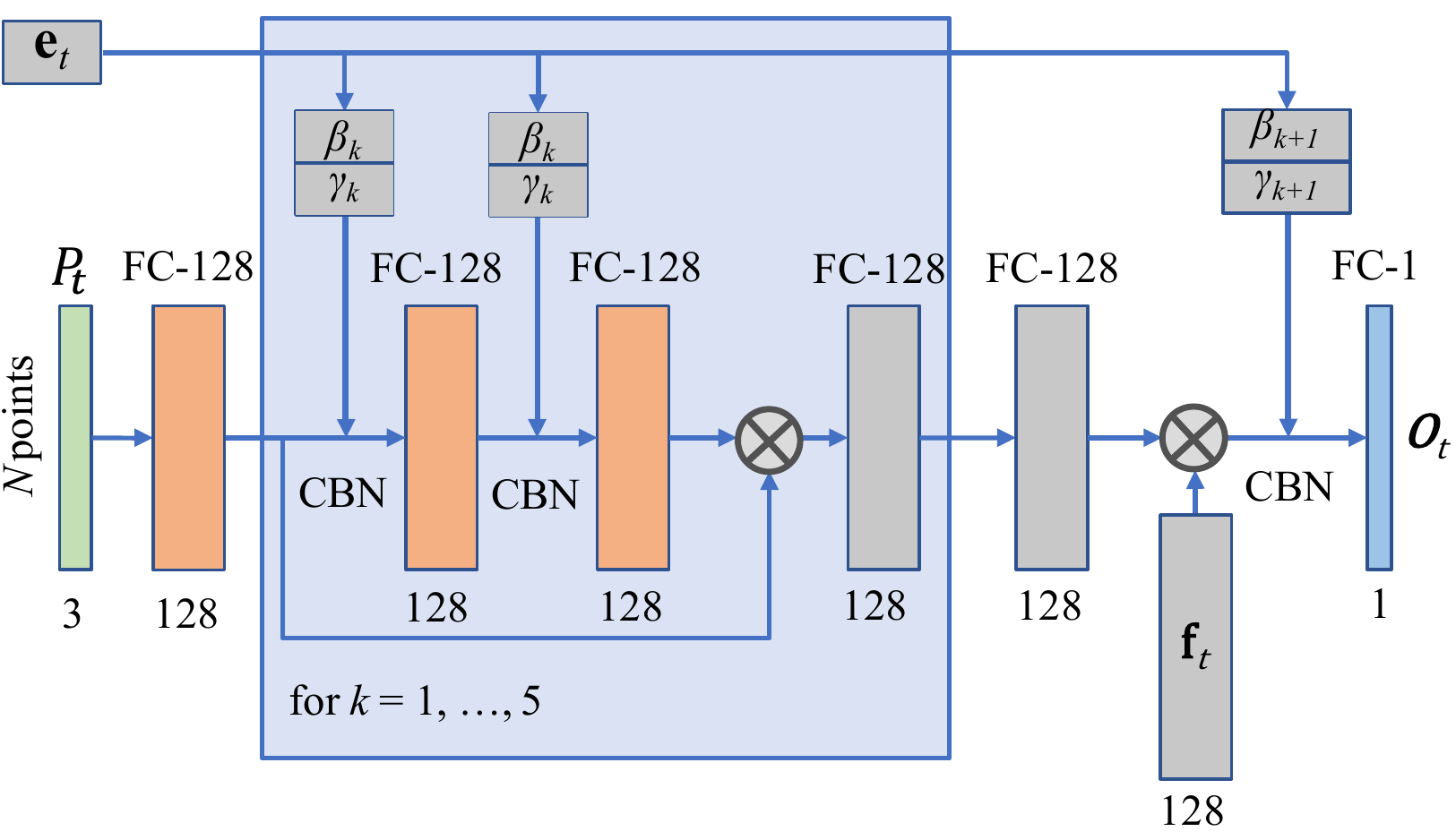}
        \caption{Occupancy Decoder}
    \end{subfigure}
    \caption{\textbf{Temporal and occupancy decoder}; $\otimes$ indicates a concatenation operation. The temporal decoder returns both a motion field $V_t$ and a motion feature map $\mathbf{f}_t$, which is then inputted to the occupancy decoder.}
    \label{fig:decoders}
\end{figure*}

\subsection{Joint Decoder} 
The joint decoder takes a spatio-temporal representation $\mathbf{e}_t$ and the original point cloud sequence as input then passes this input into two decoders (temporal decoder and occupancy decoder) to perform flow estimation and object reconstruction. Our temporal decoder and occupancy decoder are built upon the architecture from LPDC~\cite{tang2021}. However, instead of decoupling the decoders as in~\cite{tang2021}, we hypothesized
that jointly addresses two tasks by sharing information
between corresponding decoders can leverage individual
tasks. As a consequence, the close collaboration
of flow estimation and object reconstruction allows
some relaxation in the supervision need.

The temporal decoder operates as follows (see Fig.~\ref{fig:decoders} (a)). We first extract a spatio-temporal representation $\mathbf{e}_1$ for the first point cloud $P_1$ from the input sequence, using the compositional encoder. For each following point cloud $P_{t}$, we compute its spatio-temporal representation $\mathbf{e}_{t}$, then concatenate $\mathbf{e}_{t}$ with $\mathbf{e}_1$. This concatenated representation captures temporal changes of $P_{t}$ in relative to $P_1$, and is again concatenated with all points in $P_{t}$ to be processed by a series of five ResNet residual blocks~\cite{he2015deep}. Each block consists of two fully connected layers with skip connections and ReLU activation functions~\cite{Glorot2011}. The outcome of these blocks is a feature map, namely $\mathbf{f}_{t}$. This feature map is finally passed to a fully connected layer, returning a motion field $V_{t}$ describing the motion of $P_{t}$.

The occupancy decoder is slightly different from the temporal decoder (see Fig.~\ref{fig:decoders} (b)). Also different from all existing methods, our occupancy decoder works collaboratively with the temporal decoder. Particularly, input for the occupancy decoder to reconstruct the object at time step $t$ includes a point cloud $P_t$, a spatio-temporal representation $\mathbf{e}_t$ (obtained from the compositional encoder), and a flow feature map $\mathbf{f}_{t}$ (returned by the temporal decoder). The point cloud $P_t$ is first processed by a fully connected layer to extract a feature map. Similarly, the spatio-temporal representation $\mathbf{e}_t$ is fed to two different fully connected layers to obtain feature maps $\beta$ and $\gamma$. These output feature maps (from $P_t$ and $\mathbf{e}_t$) are passed to a series of five residual blocks, similar to those used in the temporal decoder. Following ONet~\cite{mescheder2019}, we apply Conditional Batch Normalization (CBN) introduced in~\cite{dumoulin2017,Vries2017} to $\beta$ and $\gamma$. Finally, the flow feature map $\mathbf{f}_{t}$ is injected into the occupancy decoder to produce an occupancy map $O_t(\mathbf{p})$, where $O_t(\mathbf{p})=1$ if the point $\mathbf{p}$ belongs to the object at time step $t$, and $O_t(\mathbf{p})=0$ otherwise.

\subsection{Joint Learning}

Like RFNet-4D~\cite{tavu2022rfnet4d}, RFNet-4D++ is trained by jointly performing two optimization processes: unsupervised learning for flow estimation and supervised learning for object reconstruction. Existing works train flow estimation using supervised learning~\cite{Fan2017,mescheder2019,Niemeyer2019,tang2021,jiang2021}, requiring fully annotated point correspondences in training data. In this paper, we propose to learn point correspondences in a point cloud sequence via an unsupervised manner, thus opening ways to new applications and more training data. Specifically, let $V_t$ be a motion field (i.e., a set of 3D vectors) at $P_t$, $V_t$ is estimated using the temporal decoder. We measure the correspondences between points in $P_t$ and $P_{t+1}$ via the Chamfer distance between $P_{t+1}$ and a translated version of $P_t$ made by $V_t$ (i.e., $P_t + V_t$). We define our flow loss as,
\begin{align}
\label{eq:motion_FW_loss}
\mathcal{L}_{\textit{flow}}=\sum_{t} \max \bigg\{&\frac{1}{|P_t|} \sum_{\mathbf{p} \in P_t + V_t} \min _{\mathbf{p}' \in P_{t+1}}\|\mathbf{p}-\mathbf{p}'\|_{2}, \nonumber \\
&\frac{1}{|P_{t+1}|} \sum_{\mathbf{p}' \in P_{t+1}} \min _{\mathbf{p} \in P_t + V_t}\|\mathbf{p}'-\mathbf{p}\|_{2} \bigg\}
\end{align}

The reconstruction task can be trained using a supervised approach. We use the conventional binary cross entropy (BCE) loss to measure the difference between predicted occupancy maps and corresponding ground truth maps. Specifically, we define our reconstruction loss as,
\begin{equation}
\label{eq:recon_loss}
\mathcal{L}_{\textit{reconstruction}}=\sum\limits_{t} \sum\limits_{\mathbf{p} \in P_{t}} \mathcal{L}_{\text{BCE}}\left(O_i\left(\mathbf{p}\right), O^{gt}_{i}\left(\mathbf{p}\right)\right)
\end{equation}
where $O_i^{gt}$ represents the ground truth occupancy map of the point cloud $P_{t}$. 

Finally, we use the following loss to train the entire architecture of RFNet-4D++,
\begin{equation}
    \label{eq:final_loss}
    \mathcal{L} = \mathcal{L}_{\textit{flow}} + \lambda \mathcal{L}_{\textit{reconstruction}}
\end{equation}
where $\lambda$ is a hyper-parameter.

To further exploit the benefit of temporal information, we train our RFNet-4D++ in both forward and backward directions in time. Particularly, we calculate the holistic temporal representation $\mathbf{h}$ for two sequences $\{P_1, ..., P_T\}$ (forward) and $\{P_T, ..., P_1\}$ (backward), and use $\mathbf{h}$ to encode the spatio-temporal representations $\mathbf{e}_t$ in both forward and backward time direction. As shown in our experiments, this training strategy improves the performance of our network in both object reconstruction and flow estimation tasks.

%%%%%%%%%%%%%%%%%%%%%%%%%%%%%% Experiments %%%%%%%%%%%%%%%%%%%%%%%%%%%%%%

\section{Experiments}
\label{sec:experiments}
\subsection{Experimental Setup}

\subsubsection{Datasets}
We experimented with our method and other baselines on the pre-processed data of D-FAUST dataset~\cite{dfaust2017}, a benchmark dataset in the field. The D-FAUST dataset contains raw-scanned and registered meshes for 129 sequences of 10 human subjects (5 females and 5 males). The objects are presented in various motions such as ``shake hips'', ``running on spot'', ``one leg jump''. We followed the train/test split used in~\cite{Niemeyer2019,tavu2022rfnet4d}. Specifically, we divided all the sequences in the D-FAUST dataset into three sets: training set (105 sequences), validation set (6 sequences), and test set (21 sequences). Since each sequence is relatively long (with more than 1,250 time steps) and in order to increase the size of the dataset, we sub-sampled each sequence into smaller sub-sequences of 17 to 50 time steps. 

We also evaluated our method on DeformingThing4D-Animals dataset~\cite{li20214dcomplete}, which consists of 1,494 non-rigidly deforming animations featuring 40 identities of 24 categories. We adopted the train/test split in~\cite{tang2022neural}. In particular, we divided all the animations into a training set (1,296 objects) and test set (198 objects). Similarly to the D-FAUST dataset, the train/test split was organised based on the identity and motion names of deforming sequences. To begin with, we initially split all the animations into two categories: seen identities and unseen identities. Within the animations of seen identities, we further separated them into two subsets: seen motions of seen identities, which we utilized as the training set, and unseen motions of seen identities, which we considered as test set S1. The animations belonging to unseen identities were then reserved exclusively for test set S2. Finally, the train, test set S1, and test set S2 consist of 1,296, 143, and 55 deforming sequences, respectively.

\begin{table*}[ht]
\centering
% \resizebox{0.8\textwidth}{!}{
\begin{tabular}{l|cccc|cccc}
\toprule
\textbf{Method} & \multicolumn{4}{c}{\stackbox{\textbf{Test set S1:} Seen Individuals, Unseen Motions}} &  \multicolumn{4}{|c}{\stackbox{\textbf{Test set S2:} Unseen Individuals, Seen Motions}} \\ \cmidrule{2-9} 
& & \textbf{IoU}$\uparrow$ & \textbf{Chamfer}$\downarrow$ ($\times 10^{-3}$) & \textbf{Corres.}$\downarrow$ ($\times 10^{-2}$) &  & \textbf{IoU}$\uparrow$ & \textbf{Chamfer}$\downarrow$ ($\times 10^{-3}$) & \textbf{Corres.}$\downarrow$ ($\times 10^{-2}$) \\ \midrule
PSGN-4D~\cite{Fan2017} &  & - &  0.6189 & 1.1083 & \multicolumn{1}{c}{} & - & 0.6877  & 1.3289 \\
ONet-4D~\cite{mescheder2019} &  & 0.7712 & 0.5921 & - & \multicolumn{1}{c}{} & 0.6827 & 0.7007 & - \\
OFlow~\cite{Niemeyer2019} &  & 0.8172 & 0.1773 & 0.8699 & \multicolumn{1}{c}{} & 0.7361 & 0.2741 & 1.0842 \\
LPDC~\cite{tang2021} &  & 0.8511 & 0.1526 & \textbf{0.7803} & \multicolumn{1}{c}{} & 0.7619 & 0.2188 & 0.9872 \\
4DCR~\cite{jiang2021} &  & 0.8171 & 0.1667 & - & \multicolumn{1}{c}{} & 0.6973 & 0.2220 & - \\
\midrule
RFNet-4D~\cite{tavu2022rfnet4d} &  & 0.8547 & 0.1504 & 0.8831 & \multicolumn{1}{c}{} & 0.8157 & 0.1594 & \textbf{0.8643} \\ 
RFNet-4D++ &  & \textbf{0.8658} & \textbf{0.1462} & 0.8709 & \multicolumn{1}{c}{} & \textbf{0.8345} & \textbf{0.1415} & 0.8855 \\ \bottomrule
\end{tabular}
% }
\caption{\textbf{Quantitative evaluation} of RFNet-4D++ and existing methods on seen and unseen individuals test splits on the D-FAUST dataset, in both reconstruction and flow estimation task. We report the volumetric IoU (higher is better), Chamfer distance (lower is better), and correspondence $\ell_2$ distance (lower is better). The notation '-' means no results, e.g., PSDN-4D does not perform reconstruction, and ONet-4D and 4DCR do not predict point correspondences across time. For each evaluation metric, the best performance is highlighted.}
\label{tab:quan_dfaust}
\end{table*}

%%%%%%%%%%%%%%%%%%%%%%%%%%%%%%%%%%%%%%%%%%%%%%%%%%%%%%%%%%%%%%%%%%%%%%%%%%%%%%%

\begin{table*}[t]
\centering
% \resizebox{0.85\textwidth}{!}{
\begin{tabular}{l|cccc|cccc}
\toprule
\textbf{Method} & \multicolumn{4}{c}{\stackbox{\textbf{Test set S1:} Seen Individuals, Unseen Motions}} &  \multicolumn{4}{|c}{\stackbox{\textbf{Test set S2:} Unseen Individuals, Seen Motions}} \\ \cmidrule{2-9} 
& & \textbf{IoU}$\uparrow$ & \textbf{Chamfer}$\downarrow$ ($\times 10^{-3}$) & \textbf{Corres.}$\downarrow$ ($\times 10^{-1}$) &  & \textbf{IoU}$\uparrow$ & \textbf{Chamfer}$\downarrow$ ($\times 10^{-3}$) & \textbf{Corres.}$\downarrow$ ($\times 10^{-1}$) \\ \midrule
LPDC~\cite{tang2021} &  & 0.7595 & 0.1531 & \textbf{0.2091} & \multicolumn{1}{c}{} & 0.5691 & 0.1321 & \textbf{0.3604} \\
RFNet-4D &  & 0.7735 & 0.1491 & 0.3049 & \multicolumn{1}{c}{} & 0.5693 & 0.1271 & 0.3871 \\
RFNet-4D++ &  & \textbf{0.7892} & \textbf{0.1342} & 0.2505 & \multicolumn{1}{c}{} & \textbf{0.5821} & \textbf{0.1204} & 0.3832 \\ \bottomrule
\end{tabular}
% }
\caption{\textbf{Quantitative evaluation} of RFNet-4D++ and existing methods on seen and unseen individuals test splits on the DeformingThing4D-Animals~\cite{li20214dcomplete} dataset, in both reconstruction and flow estimation task. We report the volumetric IoU (higher is better), Chamfer distance (lower is better), and correspondence $\ell_2$ distance (lower is better). For each evaluation metric, the best performance is highlighted. For existing methods other than RFNet-4D, LPDC is showcased here as we found it significantly outperforms other methods, e.g., PSGN-4D, ONet-4D, OFlow, 4DCR.}
\label{tab:quan_deform4d}
\end{table*}

\subsubsection{Implementation Details.} 
We implemented our method in Pytorch 1.10. We adopted Adam optimizer~\cite{adam} where the learning rate $\gamma$ was set to $10^{-4}$ and decay was set to 5,000 iterations. We empirically set $\lambda$ to 0.1 in our experiments. We trained our method with a batch size of 16, and on a single NVIDIA RTX 3090 GPU. We evaluated all the variants of our network (see Ablation Studies) on the validation set for every 2,000 iterations during the training process and used the best model of each variant on the validation set for evaluation of the variant on the test set. The training was completed once there were no further improvements achieved. For calculating the losses during training, we randomly sampled 512 points in the 3D space for the reconstruction loss, and uniformly sampled trajectories of 100 points for the flow estimation loss. More details can be found in our supplied code.

We also followed the evaluation setup used in~\cite{Niemeyer2019,tavu2022rfnet4d}. Specifically, for each evaluation, we carried out two case studies: seen individuals but unseen motions (i.e., test subjects were included in the training data but their motions were not given in the training set), and unseen individuals but seen motions (i.e., test subjects were found only in the test data but their motions were seen in the training set). 

To measure the performance of 4D reconstruction, we adopted the volumetric IoU (Intersection over Union) and the Chamfer distance reflecting the coincidence of reconstructed data and ground-truth data. To evaluate flow estimation, we used $\ell_2$-distance to measure the correspondences between estimated flows and ground-truth flows.

\subsection{Results}
We report the performance of our method on the D-FAUST dataset in Table~\ref{tab:quan_dfaust}. As shown in the results, RFNet-4D++ performs better in the seen individuals case study, for both object reconstruction and flow estimation. However, our method works consistently, and the differences in all performance metrics between the two case studies are marginal. For instance, the IoU difference between the two case studies is less than $4\%$, the differences in Chamfer distance and $\ell_2$ correspondence between these case studies are about $0.005 \times 10^{-3}$ and $0.015 \times 10^{-2}$ respectively.

In addition to the evaluation of RFNet-4D++, we also compared it with its previous version RFNet-4D~\cite{tavu2022rfnet4d} and with other existing methods including PSGN-4D~\cite{Fan2017}, ONet-4D~\cite{mescheder2019}, OFlow~\cite{Niemeyer2019}, LPDC~\cite{tang2021}, and 4DCR~\cite{jiang2021}. For the existing works, we used their published pre-trained models which had also been trained on the same training data from the D-FAUST dataset. Note that we also re-trained the previous models using their released source code. However, except for LDPC, we were not able to achieve the same results as reported in their papers. We show comparison results in Table~\ref{tab:quan_dfaust}. It can be seen that RFNet-4D++ outperforms its previous version RFNet-4D and all other baselines in 4D reconstruction using both IoU and Chamfer distance metrics. In flow estimation, RFNet-4D++ still outperforms RFNet-4D but achieves comparable performance with state-of-the-art on seen individuals, e.g., there is a slight difference (about $0.1 \times 10^{-2}$) in $\ell_2$-distance from the first ranked method (LPDC). However, unlike existing works following supervised paradigm, RFNet-4D++ is trained in unsupervised fashion requiring no point correspondence labelling. Table~\ref{tab:quan_dfaust} also show that, both RFNet-4D and RFNet-4D++ stand out in flow estimation (while RFNet-4D slightly surpasses RFNet-4D++) on unseen individual sequences, proving the ability of our architecture and joint learning of spatio-temporal representations of novel object shapes.

We report the results of RFNet-4D++, RFNet-4D, and LPDC on the DeformingThing4D-Animals dataset in Table~\ref{tab:quan_deform4d}, which show consistent performance trend with the D-FAUST dataset (i.e., reconstruction of unseen individuals with seen motions is more challenging than its counterpart case study). The results also clearly demonstrate the dominance of RFNet-4D++ over its previous version RFNet-4D and LPDC in 4D reconstruction, evident by its superior performance on both the IoU and Chamfer distance metrics. In flow estimation, RFNet-4D++ outperforms RFNet-4D on both the case studies. However, the supervised approach (LPDC) still shows its advantage over the unsupervised one (RFNet-4D, RFNet-4D++), though the difference is subtle (e.g., the averaged $\ell_2$-distance difference between the results of LPDC and RFNet-4D++ is about $0.041 \times 10^{-1}$ on the test set S1 and $0.022 \times 10^{-1}$ on the test set S2). Furthermore, labelling of point correspondences in supervised learning is extremely labour-intensive and requires complicated data capture setup. The unsupervised nature of our method opens up possibilities for broader applicability and scalability in various scenarios where labelled data may be limited or unavailable. 

%unsupervised manner does not show RFNet-4D++ achieves a performance comparable to LPDC~\cite{tang2021} for both test sets because it is trained in an unsupervised manner. This is in contrast to existing approaches that rely on a supervised paradigm, which requires annotated labels for training. Hence, our method offers a more flexible and efficient solution, as it can effectively leverage unlabelled data for learning. The unsupervised nature of our method opens up possibilities for broader applicability and scalability in various scenarios where labeled data may be limited or unavailable. 

%Fig.~\ref{fig:qualitative}. 

\begin{figure*}[!ht]
\centering
\includegraphics[width=0.96\linewidth]{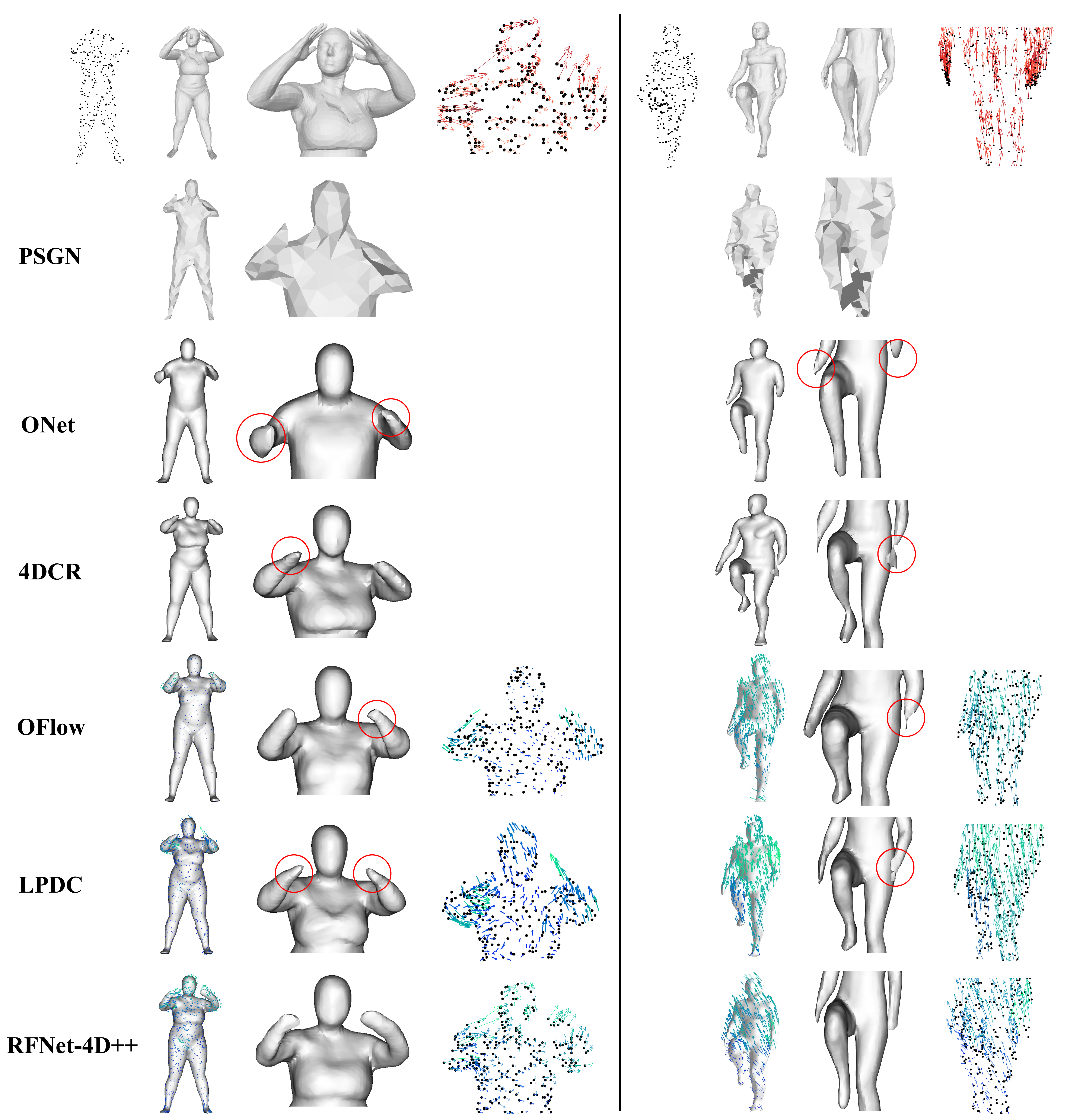}
\caption{\textbf{Qualitative evaluation} of our method and existing methods on the D-FAUST dataset. The first row includes (from left to right): input point cloud, ground truth mesh of entire body, ground truth mesh of upper/lower body, and ground-truth flows (darker vectors show stronger motions). Each following row represents corresponding reconstruction and flow estimation results. Severe errors are highlighted.} 
\label{fig:qualitative_results}
\end{figure*}

\begin{figure*}[!ht]
\centering
\includegraphics[width=0.97\linewidth]{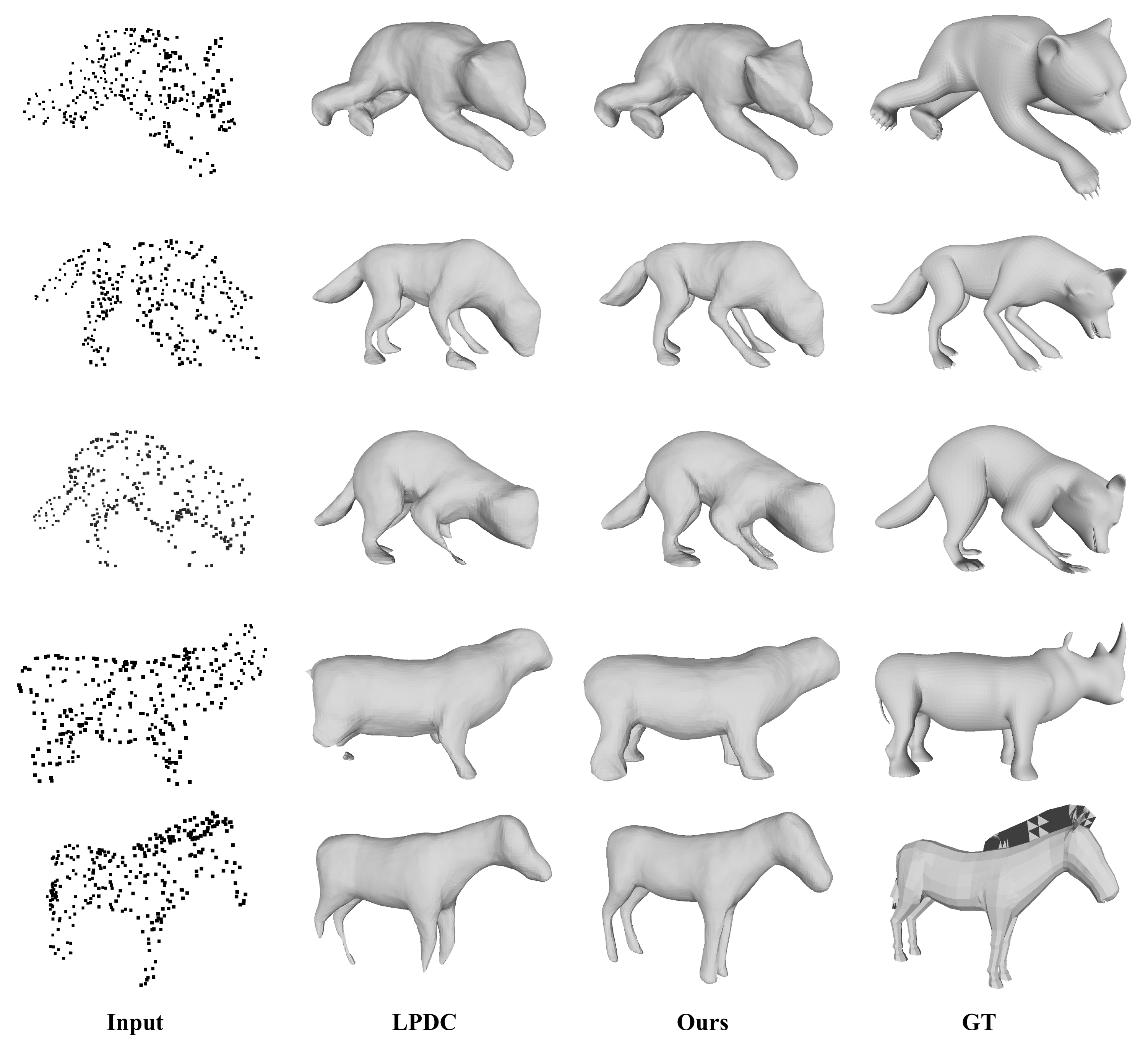}
\caption{\textbf{Qualitative evaluation} of our method and LPDC~\cite{tang2021} on the DeformingThing4D-Animals~\cite{li20214dcomplete} dataset. Note that only LPDC is showcased here as we found it significantly outperforms other methods, e.g., PSGN-4D, ONet-4D, OFlow, 4DCR.} 
\label{fig:deform4d}
\end{figure*}

We visualize several results of our RFNet-4D++ and existing methods in Fig.~\ref{fig:qualitative_results} and Fig.~\ref{fig:deform4d}. To illustrate these results, we apply the Multiresolution IsoSurface Extraction (MISE) algorithm~\cite{mescheder2019} and the Marching Cubes algorithm~\cite{marchingcubes} on reconstructed occupancy maps to generate surface meshes. As shown in the results, compared with existing methods, RFNet-4D++ achieves better reconstruction quality with more-detailed geometry recovery, e.g., the reconstructed hands produced by our method are more complete. In addition, by coupling both spatial and temporal information, the poses of body parts, e.g., the head, and the lower arms, are well preserved by our method (in reference to corresponding ground truth meshes). RFNet-4D++also shows better flow estimation than existing works as clearly demonstrated in the predicted flows in the two hands in Fig.~\ref{fig:qualitative_results}. 

%and the upper foot in Fig.~\ref{fig:qualitative}.

%In addition, we show the results of our method and LPDC~\cite{tang2021} in Figure~\ref{fig:deform4d}.

%\textcolor{blue}{To further evaluate our proposed method, we conducted additional experiments on DeformingThing4D-Animals~\cite{li20214dcomplete} dataset. The comparison result is shown in Table~\ref{tab:quan_deform4d}. The results clearly demonstrate that our method outperforms LPDC~\cite{tang2021} in terms of 4D reconstruction, as evidenced by its superior performance on both the IoU and Chamfer distance metrics. In terms of flow estimation, our method achieves a performance comparable to LPDC~\cite{tang2021} for both test sets because it is trained in an unsupervised manner. This is in contrast to existing approaches that rely on a supervised paradigm, which requires annotated labels for training. Hence, our method offers a more flexible and efficient solution, as it can effectively leverage unlabelled data for learning. The unsupervised nature of our method opens up possibilities for broader applicability and scalability in various scenarios where labeled data may be limited or unavailable. In addition, we show the results of our method and LPDC~\cite{tang2021} in Figure~\ref{fig:deform4d}.}

\begin{table*}[t]
\centering
% \resizebox{0.9\textwidth}{!}{%
\begin{tabular}{@{}c|l|ccc|ccc@{}}
\toprule
\multirow{2}{*}{\textbf{Dataset}} & \multirow{2}{*}{\textbf{Fusion}} & \multicolumn{3}{c}{\stackbox{\textbf{Test set S1:} Seen Individuals, Unseen Motions}} &  \multicolumn{3}{|c}{\stackbox{\textbf{Test set S2:} Unseen Individuals, Seen Motions}} \\ \cmidrule{3-8} 
& \textbf{technique} & \textbf{IoU}$\uparrow$ & \textbf{Chamfer}$\downarrow$ ($\times 10^{-2}$) & \textbf{Corres.}$\downarrow$ ($\times 10^{-1}$)  & \textbf{IoU}$\uparrow$ & \textbf{Chamfer}$\downarrow$ ($\times 10^{-2}$)  & \textbf{Corres.}$\downarrow$ ($\times 10^{-1}$) \\ \midrule
\multirow{3}{*}{D-FAUST}           & concat                     & 0.8547     & 0.0150        & 0.0883               & 0.8157     & 0.0159        & \textbf{0.0864}               \\
                                  & sing\_cross\_attn                & 0.8543     & 0.0147        & 0.0915               & 0.8221     & 0.0148        & 0.0933               \\
                                  & dual\_cross\_attn                & \textbf{0.8658}     & \textbf{0.0146}        & \textbf{0.0871}               & \textbf{0.8345}     & \textbf{0.0142}        & 0.0886               \\ \midrule
\multirow{3}{*}{DeformingThing4D}         & concat                     & 0.7735     & 0.1491        & 0.3049               & 0.5693     & 0.1271        & 0.3871               \\
                                  & sing\_cross\_attn                & 0.7745     & 0.1560        & 0.3288               & 0.5763     & 0.1356        & 0.3904               \\
                                  & dual\_cross\_attn                & \textbf{0.7892}     & \textbf{0.1342}        & \textbf{0.2505}               & \textbf{0.5821}     & \textbf{0.1204}        & \textbf{0.3832}               \\ \bottomrule
\end{tabular}%
% }
\caption{\textbf{Fusion techniques}. Ablation study on various fusion techniques used to fuse the spatial and temporal features in the Compositional Encoder. For each evaluation metric, the best performance is highlighted.}
\label{tab:fusion}
\end{table*}

\begin{table*}[!ht]
\centering
% \resizebox{0.9\textwidth}{!}{%
\begin{tabular}{l|l|cccccc}
\toprule
\textbf{Variant} & \textbf{Detailed setting} &  & \textbf{IoU} $\uparrow$ &  & \textbf{Chamfer} ($ \times 10^{-3}$) $\downarrow$ &  & \textbf{Corr.} ($\times 10^{-2}$) $\downarrow$ \\ \midrule
\multirow{2}{*}{Separated tasks} & Only temporal flows         &  & -      &  & -      &  & 1.5519 \\
& Only spatial points &  & 0.7712 &  & 0.5921 &  & -      \\
\midrule
Learning direction & Only FW learning      &  & 0.4988 &  & 2.4887 &  & 3.5868 \\
\midrule
\multirow{2}{*}{Flow loss} & SWD loss            &  & 0.4305 &  & 4.4621 &  & 4.0711 \\
& HD loss             &  & 0.7953 &  & 0.2103 &  & 1.3017 \\
\midrule
Flow learning & Supervised          &  & 0.8656 &  & \textbf{0.0927} &  & \textbf{0.8125} \\
\midrule
Full & Joint learning, FW-BW, Chamfer loss, unsupervised        &  & \textbf{0.8658} &  & 0.1462 &  & 0.8709 \\ \bottomrule
\end{tabular}%
% }

\caption{{\textbf{Variants of RFNet-4D++}}. Ablation study on various settings of our RFNet-4D++ on D-FAUST dataset. For each evaluation metric, the best performance is highlighted.}
\label{tab:ablation}
\end{table*}

\subsection{Ablation Studies}
%In this section, we present ablation studies to verify various aspects of the design of our model as follows: (1) fusion technique used in the Compositional Encoder, (2) variants of our proposed method, 
% (3) verification of hyper-parameter \texorpdfstring{$\lambda$}{λ} to balance the total loss, 
%(3) customisation of LPDC with RFNet-4D++'s components, and (4) impact of spatial and temporal resolutions.

In this section, we present ablation studies to verify various aspects of our method including: (1) fusion technique used in the compositional encoder, (2) variants of RFNet-4D++, 
% (3) verification of hyper-parameter \texorpdfstring{$\lambda$}{λ} to balance the total loss, 
and (3) impact of spatial and temporal resolutions.

\subsubsection{Fusion techniques}

We experimented our RFNet-4D++ with well-known fusion manners used to fuse the spatial and temporal information in the compositional encoder. Those fusion techniques include concatenation (``concat'') used in our previous RFNet-4D~\cite{tavu2022rfnet4d}, single cross-attention (``sing\_cross\_attn'')~\cite{ding2022davit}, and our proposed dual cross-attention (``dual\_cross\_attn''). Results of this experiment are presented in Table~\ref{tab:fusion}. We observed that, the single cross-attention performs on par with or slightly better than the simple concatenation operation. However, the dual cross-attention clearly and consistently boosts up the performance on both the datasets and in both the case studies (S1 and S2). This shows the ability of the dual cross-attention in learning of long-range contextual information. 

\subsubsection{Variants of RFNet-4D++}
We conducted a series of experiments on different variants of our RFNet-4D++. First, we verified our joint learning of spatio-temporal representations for 4D point cloud reconstruction and flow estimation by comparing it with the approach tackling these two tasks independently. Second, we proved the improvement of learning flows in both forward (FW) and backward (BW) directions. Third, we compared different distance metrics including the sliced Wasserstein distance (SWD) and the Hausdorff distance (HD), for the implementation of the flow loss in Eq.~(\ref{eq:motion_FW_loss}), and validated our choice, i.e., the Chamfer distance. Fourth, we experimented our model in both unsupervised and supervised fashion for flow estimation though it is intentionally designed for unsupervised learning. We summarise the results of these experiments in Table~\ref{tab:ablation}. Note that, in each experiment, only one change was applied at a time while other settings remained unchanged. For settings using either temporal or spatial information (see the first two rows in Table~\ref{tab:ablation}), only the corresponding encoder and decoder (i.e., spatial/temporal encoder and decoder) were activated while the counterpart encoder and decoder were frozen. These settings correspond to solving the flow estimation and reconstruction tasks separately. To experiment our method with supervised learning for flow estimation, we followed the settings used in OFlow~\cite{Niemeyer2019}. In particular, we used ground-truth point correspondences from the training data and $\ell_2$ distance for the motion loss, i.e., replacing the Chamfer distance in Eq.~(\ref{eq:motion_FW_loss}) by $\ell_2$ distance. Note that, the D-FAUST dataset is fully annotated with point correspondences and thus also supports supervised learning. When training our model in an unsupervised manner, those point correspondences were not used. Experimental results in Table~\ref{tab:ablation} clearly confirm the design of our RFNet-4D++ in both object reconstruction and flow estimation.

\begin{figure*}[!ht]
  \centering
  \includegraphics[width=0.32\linewidth]{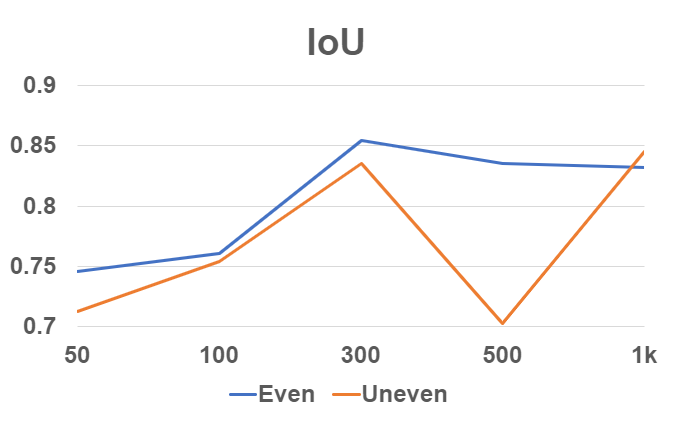}
  \includegraphics[width=0.32\linewidth]{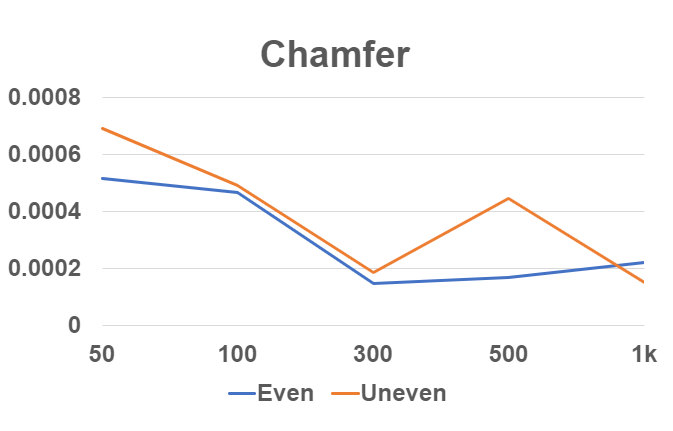}
  \includegraphics[width=0.32\linewidth]{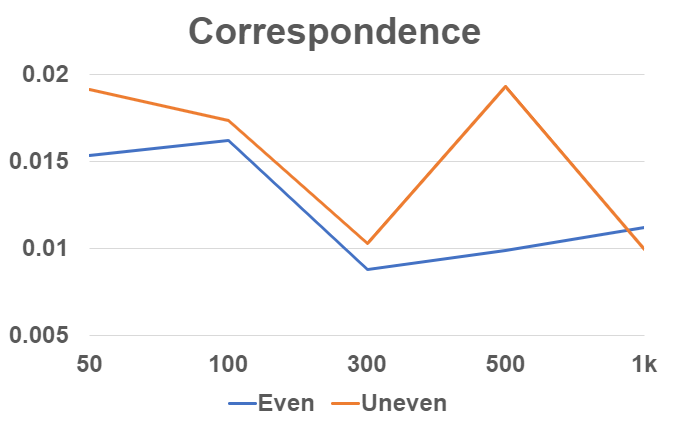}

  \includegraphics[width=0.32\linewidth]{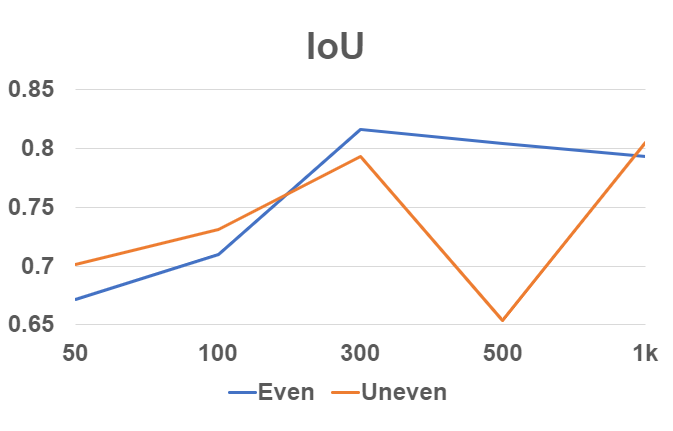}
  \includegraphics[width=0.32\linewidth]{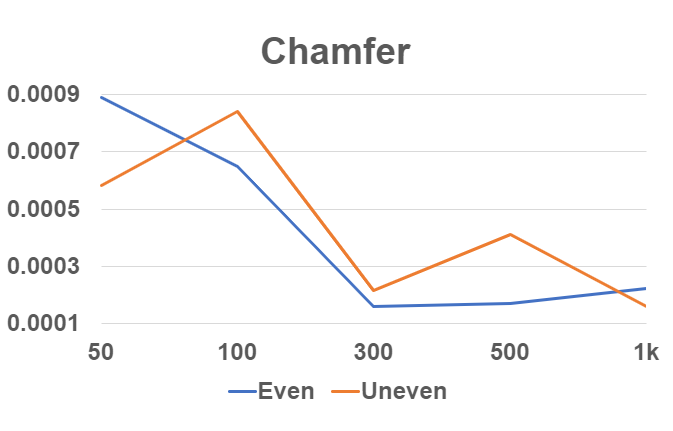}
  \includegraphics[width=0.32\linewidth]{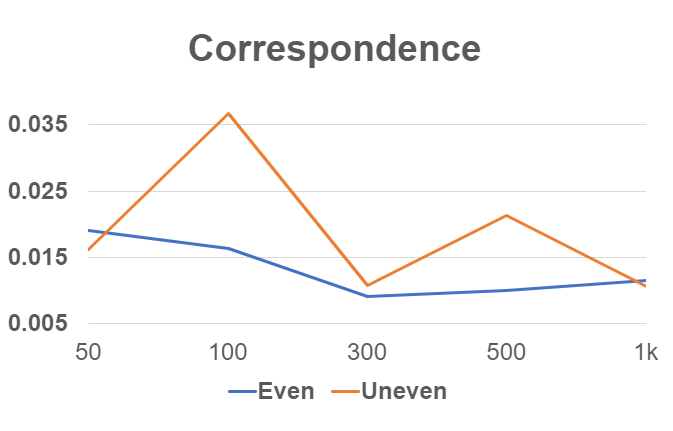}
  
  \caption{RFNet-4D++ under various spatial/temporal resolutions on both seen (top row) and unseen (bottom row) individuals test splits on the D-FAUST dataset in both reconstruction and flows estimation tasks.}
  \label{fig:spatio-temporal-resolutions-full}
\end{figure*}

\subsubsection{Impact of spatial and temporal resolutions} 

We tested our method under various spatial and temporal resolutions on both seen and unseen individuals test splits, and in both reconstruction and flow estimation tasks. For spatial resolutions, we randomly sampled an input point cloud with 50, 100, 300 (the input used in all experiments), 500, and 1k point trajectories from ground-truth surfaces. For each spatial resolution, like LPDC, we varied the temporal resolutions by sampling each point cloud sequence in two ways: even sampling (as used in all experiments) and uneven sampling (where we randomly selected 6 frames with large variations between adjacent frames from a 50-time step segment). We present the results of this experiment in Fig.~\ref{fig:spatio-temporal-resolutions-full}.

The relationship between input point density and reconstruction and flow estimation performance is clearly shown in Fig.~\ref{fig:spatio-temporal-resolutions-full}. In general, increasing the number of input points within the range of 50-300 points leads to improved performance. However, when the input density exceeds 300 points (which was the case in all experiments), a noticeable drop in performance occurs. This decline can be attributed to the complexities involved in establishing flow correspondences for denser point clouds. Despite this decrease, the performance of denser input points still outperforms sparser input points, indicating their relative advantage even in challenging scenarios.
For the temporal resolution-related experiment, it is worth noting that even or uniformly sampled input points consistently yield superior performance compared to uneven sampling. This is likely due to the fact that point locations and flow correspondences can undergo significant changes or exhibit large variations, making flow estimation more challenging when unevenly distributed points are used.

\subsection{Complexity Analysis}
We provide a complexity analysis on the memory footprint and computational efficiency of our RFNet-4D++ and its previous version RFNet-4D,  and several existing models including OFlow~\cite{Niemeyer2019} and LPDC~\cite{tang2021} (current state-of-the-art). In this experiment, we trained all the models with a batch size of 16, using a sequence of 17-time steps with consistent intervals. All the models were run on a single NVIDIA RTX 3090. We report the memory footprint, training, and inference time in Table~\ref{tab:time}. For the training time, we computed the average of batch training time throughout the first 100k iterations of training (seconds per iteration). For the inference time, we reported the average time required to infer using a batch size of 1 for 1k test sequences (seconds per sequence). As shown in Table~\ref{tab:time}, despite our model taking a larger memory footprint for training, its training time is approximately 3.5 times and 1.6 times faster than that of OFlow and LPDC respectively. Similarly, our model performs 1.9 times and 4 times faster than OFlow and LPDC in inference. We found that OFlow takes a much longer time for training since OFlow makes use of an ODE-solver requiring intensive computations and gradually increasing the number of iterations to fulfill error tolerance.

\begin{table}[t]
\centering
\begin{tabular}{l|ccc}
\toprule
\textbf{Method} & \textbf{Memory} & \textbf{Training} (sec/iter) & \textbf{Inference} (sec/seq) \\ 
% \multirow{2}{*}{\textbf{Method}} & \multirow{2}{*}{\textbf{Memory}} & \textbf{Training} & \textbf{Inference} \\ 
 % &  & (sec/iter) & (sec/seq) \\ 
\midrule
OFlow~\cite{Niemeyer2019} & 3.96GB & 4.65s & 0.95s \\
LPDC~\cite{tang2021} & 11.90GB & 2.09s & 0.44s \\
RFNet-4D & 14.20GB & 1.33s & 0.24s \\ 
RFNet-4D++ & 16.87GB & 1.54s & 0.29s \\ 
\bottomrule
\end{tabular}
\caption{\textbf{Space and time complexity} of our method and existing ones on the D-FAUST dataset.}
\label{tab:time}
\end{table}

%%%%%%%%%%%%%%%%%%%%%%%%%%%%%% Conclusion %%%%%%%%%%%%%%%%%%%%%%%%%%%%%%
\section{Discussion and Conclusion}
\label{sec:conclusion}

This paper proposes RFNet-4D++, a network architecture for jointly reconstruction of objects and estimation of temporal flows from dynamic point clouds. The proposed network is built upon a compositional encoder effectively capturing informative spatio-temporal representations for 4D point clouds, and a joint learning paradigm leveraging sub-tasks to improve overall performance. We extensively evaluated our method and compared it with existing works on two benchmark datasets including D-FAUST and DeformingThing4D datasets. Experimental results demonstrated the effectiveness and efficiency of our method in comparison with the current state-of-the-art.

There is also room for future research. First, we found that existing 4D reconstruction methods often suffer from large displacements between data frames. Second, their reconstruction quality tends to drop over time due to accumulated errors. It is also worthwhile to study 4D reconstruction for different types of objects and with more challenging input data types, e.g., LiDAR point clouds that are commonly used in autonomous driving applications.

\section{Acknowledgement} 
This paper was partially supported by an internal grant from HKUST (R9429).

\appendices
\section{Occupancy Decoder}
\label{sec:occupancy_decoder}

We depict our occupancy decoder in Fig.~\ref{fig:decoders}(b). 
% Note that this figure has also been provided in the paper and is replicated in this supplementary material for convenience in description. 
The occupancy decoder takes a spatio-temporal representation $\mathbf{e}_{t}$ from the compositional encoder, a point cloud including $N$ points, and a motion feature map $\mathbf{f}_{t}$ from the temporal encoder as input. The compositional encoder and temporal decoder are described in Section~\ref{sec:proposed_method}. The points are passed through a fully-connected layer to produce a 128-dimensional feature vector for each point. This feature vector is then passed through five (5) pre-activation ResNet blocks. Each ResNet block first applies Conditional Batch-Normalization (CBN)~\cite{dumoulin2017,Vries2017} to the current feature vector followed by a ReLU activation function. The output is then fed into a fully-connected layer, a second CBN layer, a ReLU activation, and another fully-connected layer. The output of this operation is then added to the input of the ResNet block. The output, after going through five ResNet-blocks, is concatenated with $\mathbf{f}_{t}$ before being finally passed through the last CBN layer and ReLU activation followed by a fully-connected layer to project the features down to one dimension. To obtain an occupancy map for an input point cloud, this vector can simply be passed through a sigmoid activation.

Following~\cite{BN2015}, the CBN layers are implemented in the following way. First, we pass $\mathbf{e}_{t}$ through two fully-connected layers to obtain 128-dimensional learnable parameter vectors $\beta(\mathbf{e}_{t})$ and $\gamma(\mathbf{e}_{t})$. We next normalize a 128-dimensional input feature vector $f_{in}^{BN}$ using first and second-order moments, then multiply the normalized output with $\gamma(\mathbf{e}_{t})$ and finally add the bias term $\beta(\mathbf{e}_{t})$ as, 
$$
f_{out}^{BN}=\gamma(\mathbf{e}_{t}) \frac{f_{in}^{BN}-\mu}{\sqrt{\sigma^{2}+\epsilon}}+\beta(\mathbf{e}_{t}),
$$
where $\mu$ and $\sigma$ are the empirical mean and standard deviation (over the batch) of the input features $f_{in}^{BN}$ and $\epsilon=10^{-5}$ (the default value of PyTorch). Moreover, we compute a running mean over $\mu$ and $\sigma^{2}$ with momentum $0.1$ during training. At inference time, we replace $\mu$ and $\sigma^{2}$ with the corresponding running mean.

\section{Evaluation Details}
\label{sec:evaluation_details}

We use the volumetric IoU and Chamfer distance for evaluation of object reconstruction at each time step. Following~\cite{mescheder2019}, for each point cloud, we randomly sample 100,000 points inside or on the reconstructed result (i.e., predicted mesh) of the point cloud and determine whether these points lie inside or outside its corresponding ground truth mesh. Then the volumetric IoU is computed as the ratio of the volume of the intersection and union of the predicted mesh and ground truth mesh. The Chamfer distance is defined as the mean of accuracy and a completeness metric. The accuracy metric is defined as the mean distance of points on the predicted mesh to their nearest neighbors on the ground truth mesh. The completeness metric is defined similarly, but in the opposite direction, i.e., given a ground truth point, its nearest neighbor on the predicted mesh is used. We estimate both accuracy and completeness metrics efficiently by randomly sampling 100,000 points from both predicted and ground truth mesh and using a KD-tree to estimate corresponding nearest neighbors from the meshes. 

To evaluate flow estimation, we adopt the $\ell_2$-distance used in~\cite{Niemeyer2019}. Specifically, given a reconstructed mesh and a 3D vector field at a time step $t$ of a point cloud $P_t$, we apply the vector field on vertices of the reconstructed mesh to estimate their locations at time step $t+1$. We then find the corresponding nearest points of these new locations on the ground truth mesh of $P_{t+1}$ at time step $t+1$. These nearest points can be computed efficiently using a KD-tree. We finally measure the mean of $\ell_2$-distances between estimated locations and their corresponding nearest points. Note that point correspondences are available in the D-FAUST dataset. However, since our RFNet-4D is trained to estimate motion flows in an unsupervised manner, these point correspondences are only used for evaluation. In contrast, they are used for both training and evaluation in existing methods, e.g., PSGN-4D, OFlow, and LPDC.

%\begin{figure*}[!t]
%  \centering
%  \includegraphics[width=0.96\linewidth]{figures/Qualitative_Results_final_3-4.pdf}
%  \caption{\textcolor{blue}{\textbf{Qualitative results} of our method and existing methods on D-FAUST dataset. The first row includes (from left to right): input point cloud, ground truth mesh of entire body, ground truth mesh of lower body, and ground truth flows (darker vectors show stronger motions). Each following row represents corresponding reconstruction and flow estimation results. Severe errors are highlighted.}}
%  \label{fig:qualitative}
%\end{figure*}

\ifCLASSOPTIONcaptionsoff
  \newpage
\fi

\bibliographystyle{IEEEtran}
\bibliography{references}



% \end{thebibliography}

% \newpage
\begin{IEEEbiography}
[{\includegraphics[width=1in,height=1.25in,clip,keepaspectratio]{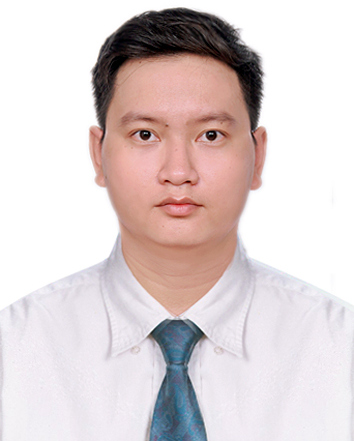}}]{Tuan-Anh Vu} is a final year Ph.D. student at Hong Kong University of Science and Technology (HKUST). He received a Bachelor's degree from International University, Vietnam National University, Ho Chi Minh City, Vietnam in 2016. His current research interests include Deep Learning, Unsupervised Learning, 3D Reconstruction, Scene Understanding, and Shape Analysis.
\end{IEEEbiography}
% \vskip -2\baselineskip plus -1fil
\begin{IEEEbiography}
[{\includegraphics[width=1in,height=1.25in,clip,keepaspectratio]{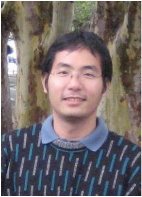}}]{Duc Thanh Nguyen} is a Senior Lecturer within the School of Information Technology, Deakin University, Victoria, Australia. Nguyen's research interests include Computer Vision, Machine Learning, and Multimedia Computing. He has published his work in highly-ranked publication venues in the field such as Pattern Recognition journal, IEEE Transactions, CVPR, ICCV, ECCV, KDD, and AAAI. He has been an Area Chair of the Multimedia Analysis and Understanding track for the IEEE International Conference on Multimedia and Expo since 2021. He has been a Reviewer for many international journals and a Technical Program Committee Member for many premium conferences in his research field. Nguyen has attracted and managed competitive national/international research funding with a total income over \$2.5 mil AUD.
\end{IEEEbiography}
% \vskip -2\baselineskip plus -1fil
\begin{IEEEbiography}[{\includegraphics[width=1in,height=1.25in,clip,keepaspectratio]{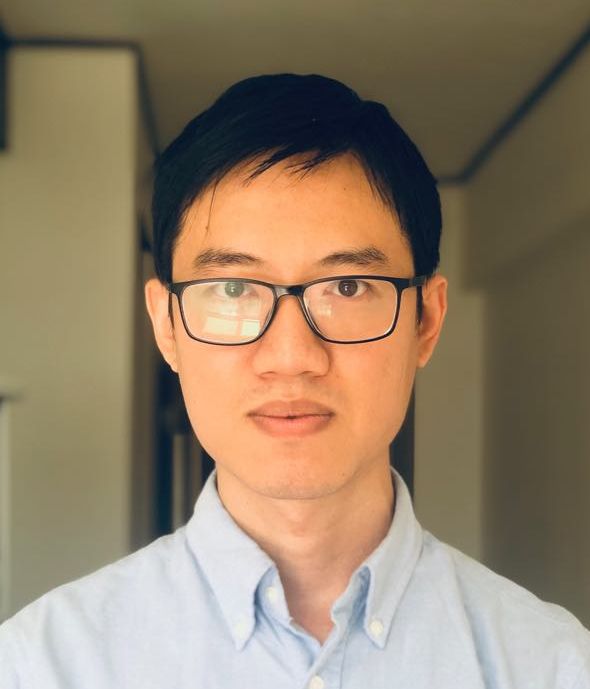}}]{Binh-Son Hua}
is currently an Assistant Professor with the School of Computer Science and Statistics, Trinity College Dublin. His research interests are at the intersection of computer graphics, computer vision, and machine learning with a focus on 3D deep learning and realistic image synthesis. He received his Ph.D. in Computer Science from National University of Singapore in 2015, and spent his postdoctoral research at Singapore University of Technology and Design and The University of Tokyo. He received the Best Paper Honorable Mention at 3DV 2016. 
\end{IEEEbiography}
% \vskip -2\baselineskip plus -1fil
\begin{IEEEbiography}[{\includegraphics[width=1in,height=1.25in,clip,keepaspectratio]{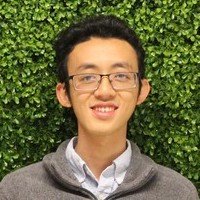}}]{Quang-Hieu Pham} is a Software Engineer at Woven by Toyota, USA. He received his Ph.D. in Computer Science from Singapore University of Technology and Design (SUTD) in 2020, advised by Prof. Sai-Kit Yeung and Prof. Gemma Roig. He received his Bachelor's degree at Ho Chi Minh University of Science in 2014. His main research interests are 3D computer vision and deep learning.
\end{IEEEbiography}
% \vskip -2\baselineskip plus -1fil
\begin{IEEEbiography}[{\includegraphics[width=1in,height=1.25in,clip,keepaspectratio]{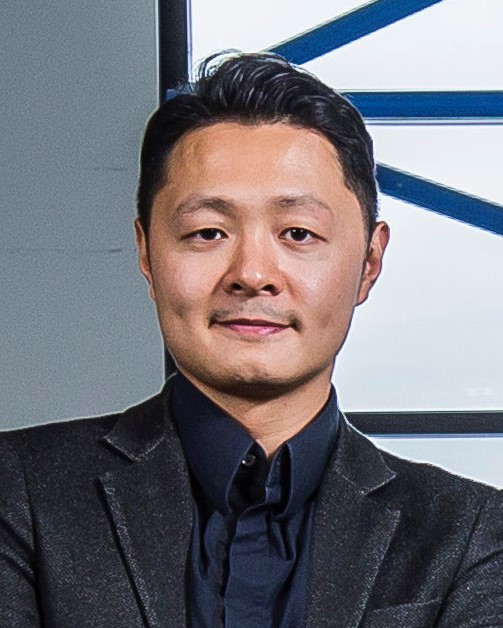}}]{Sai-Kit Yeung} is an Associate Professor at the Division of Integrative Systems and Design (ISD) and the Department of Computer Science and Engineering (CSE) at the Hong Kong University of Science and Technology (HKUST). His research interests include 3D vision and graphics, content generation, fabrication, novel computational techniques, and integrative systems for marine-related problems. He has served as a Senior Program Committee member in IJCAI and AAAI and as a Course Chair for SIGGRAPH Asia 2019. In addition, he regularly serves as a Technical Papers Committee member for SIGGPAPH \& SIGGRAPH Asia and is currently an Associate Editor of the ACM Transactions on Graphics (TOG).
\end{IEEEbiography}


\vfill


\end{document}